%% file: main.tex
\documentclass[10pt, a4paper]{article}
\usepackage{amsmath}

\usepackage{amssymb}% http://ctan.org/pkg/amssymb
\usepackage{pifont}% http://ctan.org/pkg/pifont
\newcommand{\cmark}{{\color{teal}\ding{51}}}%
\newcommand{\xmark}{{\color{red}\ding{55}}}%
\usepackage[utf8]{inputenc}
\usepackage{caption}
\usepackage{subfigure}

\usepackage{scalerel,xparse}

\usepackage[frozencache,cachedir=.]{minted}
\usepackage{booktabs}

\usepackage{microtype}
\usepackage{etoolbox}
\newcommand{\emojiro}{\scalerel*{\includegraphics{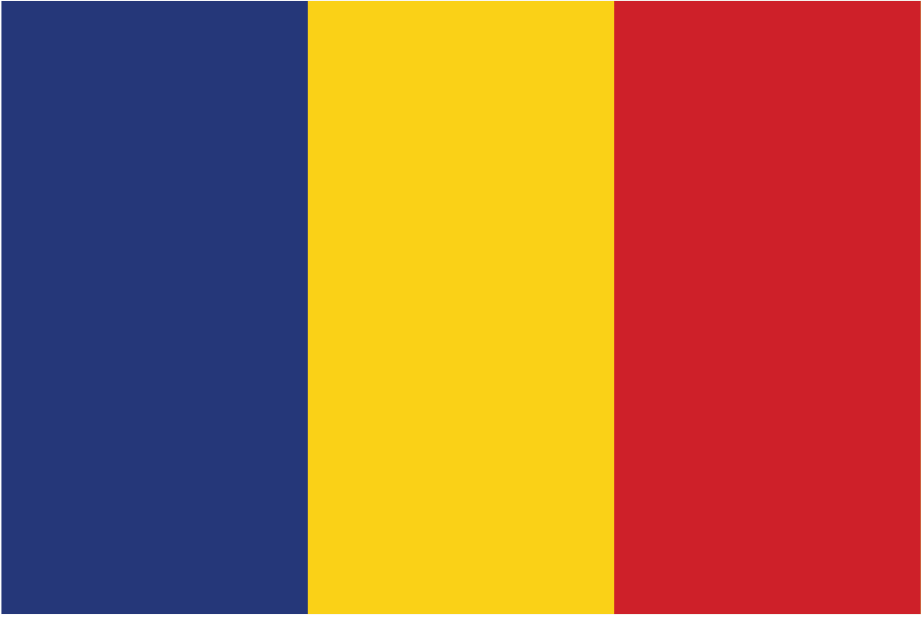}}{X}}

\usepackage[final]{lrec-coling2024} % this is the new style

\title{RoCode: A Dataset for Measuring Code Intelligence from Problem Definitions in \emojiro \hspace{0.01cm} Romanian \emojiro}
\name{Adrian Cosma$^1$, Bogdan Iordache$^2$, Paolo Rosso$^{3,4}$} 

\address{$^1$University POLITEHNICA of Bucharest\\ 
$^2$ University of Bucharest, HTL Research Center\\
$^3$ PRHLT Research Center, Universitat Politècnica de València\\
$^4$ ValgrAI - Valencian Graduate School and Research Network of Artificial Intelligence\\
$^{1,2}$Bucharest, Romania,  $^{3,4}$València, Spain\\
ioan\_adrian.cosma@upb.ro, iordache.bogdan1998@gmail.com, prosso@dsic.upv.es\\}

\abstract{
Recently, large language models (LLMs) have become increasingly powerful and have become capable of solving a plethora of tasks through proper instructions in natural language. However, the vast majority of testing suites assume that the instructions are written in English, the de facto prompting language. Code intelligence and problem solving still remain a difficult task, even for the most advanced LLMs. Currently, there are no datasets to measure the generalization power for code-generation models in a language other than English. In this work, we present RoCode, a competitive programming dataset, consisting of 2,642 problems written in Romanian, 11k solutions in C, C++ and Python and comprehensive testing suites for each problem. The purpose of RoCode is to provide a benchmark for evaluating the code intelligence of language models trained on Romanian / multilingual text as well as a fine-tuning set for pretrained Romanian models. Through our results and review of related works, we argue for the need to develop code models for languages other than English.
 \\ \newline  }

\begin{document}

\maketitleabstract

\nocite{*}

% https://colorhunt.co/palette/219c90e9b824ee9322d83f31

\section{Introduction}
\input{sections/1.introduction}

\section{Related Work}
\input{sections/2.related}

\section{RoCode: Romanian Competitive Programming}
\input{sections/3.method}

\section{Results}
\input{sections/4.results}

\section{Conclusions}
\input{sections/5.conclusions}

\section*{Acknowledgements}
\input{sections/6.ack}

\section{References}\label{sec:reference}
\bibliographystyle{lrec-coling2024-natbib}
\bibliography{refs}

\end{document}

%% file: sections/1.introduction.tex
Since the development of large language models (LLMs) \cite{NEURIPS2020_1457c0d6,hoffmann2022training,borgeaud2021improving,palm-lm}, few tasks have been left behind that cannot be reasonably tackled with proper prompting \cite{guo2023evaluating}. Of particular interest in this area has been natural language to programming language (NL-PL) capabilities, in which models generate structured code with a precise intent \cite{replit-code,luo2023wizardcoder,roziere2023code}. LLMs have been particularly successful in this area, fueled by the massive amounts of available code on the internet \cite{kocetkov2022stack}.

"Low-code" platforms, which enable users to develop software requiring less coding knowledge, require efficient interfacing between human operators and machine code. One of the most prominent tools in this direction is Github Copilot \cite{chen2021evaluating}, a large language model trained to generate code based on natural language comments. In recent years, a wide array of methods that improve upon the state of the art of code execution have been proposed \cite{scholak2021picard,christopoulou2022pangu,ni2023lever,zhou2023large}, including prompt manipulation methods such as chain-of-thought \cite{wei2022chain} or similar approaches \cite{zhou2023large} which appear to elicit reasoning capabilities. The most powerful commercial language model, GPT-4 \cite{openai2023gpt4}, achieves great performance in a wide array of tasks, but its performance is still lacking in programming puzzles. While there is no information about the pretraining dataset composition, nor the composition of benchmarks, GPT-4's results on a benchmark dubbed “Leetcode” appears to be the worst performing, especially the hard subset, correctly solving only 3 out of the 45 problems. It is unclear how much the performance of GPT-4 on programming problems is due to a high degree of generalization, or due to data leakage from other websites such as LeetCode\footnote{\url{https://leetcode.com/}}.  There are several existing datasets for semantic code search and competitive programming \cite{li2022competition,chen2021evaluating,husain2019codesearchnet,iyer2018mapping,zavershynskyi2018naps,spoc}, but almost all of them have problem statements and comments written in English. Furthermore, out of the currently available open-sourced large language models (e.g., LLaMa \cite{touvron2023llama,roziere2023code}), the vastly predominant pretraining language is English. 

By design, low-code systems promise the democratization of programming. In itself, coding is independent of the native language of the programmer. However, most NL-powered low-code platforms have a tacit requirement that the user is fluent in English.

\input{table/dataset-comparison}

Even through tremendous progress, multilingual systems \cite{DBLP:journals/corr/abs-2002-10957,bloom} lag behind specialized models, especially in languages with fewer training resources available \cite{bloom}. For instance, RoBERT \cite{dumitrescu-etal-2020-birth}, the Romanian version of the popular BERT \cite{devlin-etal-2019-bert} outperforms multilingual counterparts on Romanian tasks. Moreover, translation is imperfect due to the lexical gap between languages which makes some concepts to be difficult to translate directly and can induce a loss of meaning that might be crucial in certain high-stakes scenarios. The current state of the art for translating Romanian to English \cite{bhosale2020language} has a reported 40.3 BLEU score, which is considered "understandable to good translation", leaving a lot to be desired in meaning-rich contexts such as code generation. Other methods such as NLLB \cite{nllb2022} provide higher quality translations, but are not open-source.

Moreover, code from non-English speaking countries often exhibits \textit{code-code-switching}: programming syntax keywords are written in English, while comments and domain-specific attributes (i.e. variables, class names) are written in the native language. This phenomenon is, in fact, considered a best practice and falls in line with the idea of "ubiquitous language" \cite{evans2004ddd}: domain experts and developers need to share a single, common vocabulary such that the meaning is exact and not lost in translation.

Efforts to bring the power of natural language-powered systems to other languages apart from English are limited. For Romanian, only the RoGPT-2 \cite{9643330}, GPTNeo-Ro\cite{ro-transformers}, and RoBERT \cite{dumitrescu-etal-2020-birth} counterparts are available. These models achieved good performance on LiRo \cite{liro2021}, the current Romanian benchmarking suite, compared to other similar multilingual models. However, there is no benchmark for evaluating code generation models for Romanian. Furthermore, small scale models ($<1$B parameters) fare poorly on coding challenges \cite{hendrycksapps2021}. Nevertheless, a feasible alternative is the construction of code understanding / code retrieval models adapted for Romanian, such as CodeBERT \cite{feng2020codebert}.

In this work, we propose RoCode, the first competitive programming dataset for measuring code intelligence for NL-PL models. RoCode consists of 2,642 problems written in Romanian under 3 difficulty levels, multiple associated solutions written in C / C++ and Python, alongside a set of test cases to evaluate the correctness and algorithmic complexity. RoCode attempts to bridge the gap between Romanian natural language and computer code. RoCode is the first dataset of competitive coding problems in a language different from English. Problems, solutions and test cases are made available through a collaboration with \textit{\url{infoarena.ro}}, the most popular Romanian competitive programming platform. While problems and solutions are publicly available to be crawled, the test cases for each problem are not. We provide a filtered, curated and structured dataset, containing test cases for each problem, as well as an open-source environment to test generated solutions.

Compared to other existing datasets for competitive programming, such as APPS \cite{hendrycksapps2021}, RoCode is similar in size and scope, while having its own particularities geared towards Romanian. RoCode has problem definitions written in Romanian, and solutions exhibit \textit{code-code-switching}, creating a challenging set for fine-tuning monolingual models. In Table \ref{tab:dataset-comparison} we provide a comparison with other similar datasets \cite{iyer2018mapping,zavershynskyi2018naps,spoc}. Through RoCode, we aim to facilitate the development of NL-PL models in native Romanian, outperforming current multilingual models. RoCode aims to be a benchmark in neural code generation from Romanian prompts as well as a fine-tuning dataset for larger models.

This work makes the following contributions:
\begin{enumerate}
    \item We propose RoCode, the first dataset for measuring code intelligence from problem definitions written in Romanian. We provide 2,642 problems under 3 difficulty levels, solutions in C / C++ and Python, and test cases for each problem. We release the dataset on \textit{huggingface} for public use\footnote{\url{huggingface.co/datasets/cosmadrian/rocode}}.
    
    \item We provide the first results on NL-PL code intelligence performance on small open-source models on a language different than English. We tested all the available Romanian language models (RoGPT-2 \cite{9643330} and GPT-Neo-Ro \cite{ro-transformers}) and a set of open-source English models \cite{replit-code,openlm2023openllama,luo2023wizardcoder,touvron2023llama}. Unsurprisingly, none of the tested models are able to obtain a reasonable performance, proving that RoCode is a challenging dataset. We make our code publicly available on github\footnote{\url{github.com/cosmaadrian/rocode}}.
    
    \item At the same time, this work is also a position paper. Through our results and extensive review of related works, we argue for the development of multi-lingual and monolingual non-English code intelligence models and provide potential future research directions.
\end{enumerate}

%% file: table/dataset-comparison.tex
\begin{table*}[hbt]
    \centering
    \resizebox{\linewidth}{!}{
        \begin{tabular}{l|cccccc}
              & CONCODE & NAPS & SPoC & APPS & \textbf{RoCode (ours)} \\
             \midrule
            Programming Language & Java & UAST & C++ & Python &\textbf{ C / C++ / Python} \\
            Test Cases &  \xmark  & \cmark & \cmark  &  \cmark & \cmark  \\
            Number of Programs & 104,000 & 17,477 & 18,356 & 232,421 &  \textbf{11,250} \\
            Lines per Program (Avg.)  & 26.3 & 21.7 & 14.7 & 18.0 & \textbf{118.65} \\
            Number of Exercices  & 104,000 & 2,231 & 677 &  10,000 & \textbf{2,642} \\
            Text Input & Docstrings & Pseudocode &  Pseudocode & Problem Descriptions & \textbf{\emojiro \hspace{0.05cm} Problem Descriptions} \\
        \end{tabular}
    }
    \caption{Comparison with other existing NL-PL datasets. While RoCode has a comparable number of problems and solutions, its problem descriptions are formulated in native Romanian. Furthermore, solutions are written by Romanian students and can exhibit ``code-code-switching''.}
    \label{tab:dataset-comparison}
\end{table*}

%% file: sections/2.related.tex
\subsection{Large Language Models for Code}

Following the success of ChatGPT models to generate and understand code, multiple other very recent open-source alternatives have been proposed \cite{bloom,roziere2023code,fried2023incoder,luo2023wizardcoder,li2023starcoder} that make use of publicly available data (e.g., The Stack dataset \cite{kocetkov2022stack}), as well as synthetically generated data \cite{wang2023selfinstruct}. For instance, StarCoder \cite{li2023starcoder} was trained on publicly available data and obtained impressive results on code benchmarks, surpassing in some cases, proprietary models. WizardCoder \cite{luo2023wizardcoder} is a subsequent improvement through instruction fine-tuning of StarCoder with the Evol-Instruct method.

Notably, CodeLlama \cite{roziere2023code}, a LLaMa-derived \cite{touvron2023llama} model, has received increased attention due to its high quality, multiple model sizes and its permissive open-source language. CodeLLaMa is trained on 500B tokens of publicly available code. Several specialized variants are released, including a model tuned specifically for Python and an instruction-following model. The model is trained using infilling \cite{bavarian2022efficient} (similar to InCoder \cite{fried2023incoder}) and automatically generated instructions from a larger model using the self-instruct method \cite{wang2023selfinstruct}.

However, almost all open-sourced LLMs are primarily geared towards English (only BLOOM-176B \cite{bloom} is multilingual), and all of the code-focused LLMs are exclusively fine-tuned on code having comments and documentation written in English \cite{pile,kocetkov2022stack}.

\subsection{Code Datasets}
Large publicly available training datasets for code intelligence have made heavy use of codebases hosted on GitHub with permissive licenses \cite{pile,kocetkov2022stack,husain2019codesearchnet}. More notably, the Pile \cite{pile} is an open 800GB dataset of text having a considerable fraction comprised of code in various languages. Similarly, the Stack \cite{kocetkov2022stack} is a 3TB dataset of code from 30 programming languages, used to train StarCoder \cite{li2023starcoder}. Some works such as Codex \cite{chen2021evaluating} use GitHub to compile a dataset, but do not disclose the repository details or licensing.

For benchmarking problem solving capabilities of LLMs, one predominant dataset used across approaches is APPS \cite{hendrycksapps2021}, a dataset of leetcode-style problems organised into difficulty ranges. The APPS dataset withstood the test of time and has proven to be a hard benchmark even for the most sophisticated models. Other benchmarking datasets have been proposed in the past \cite{iyer2018mapping,zavershynskyi2018naps,spoc}, but they are geared towards code generation from some text input (docstrings or pseudocode) rather than solving a specific programming problem.

%% file: sections/3.method.tex
For the data collection, we collaborated with InfoArena\footnote{\url{infoarena.ro}}, one of the most popular Romanian competitive programming websites. The platform hosts a total of 3,072 coding problems, with difficulty ranging from simple to high school International Olympiad level. The problems have a problem description written in Romanian, alongside descriptions of input and output requirements and several easy test cases for users to evaluate the solution. Users can submit solutions written in C / C++, which are automatically evaluated in a sandbox environment. Solutions are stored alongside the number of passed test cases. Test cases evaluate both the logical correctness of the solutions and their algorithmic complexity (i.e., sub-optimal solutions are given a lower score). 

\subsection{Problem Statements}
\begin{figure*}[hbt!]
    \centering
    \includegraphics[width=1.0\textwidth]{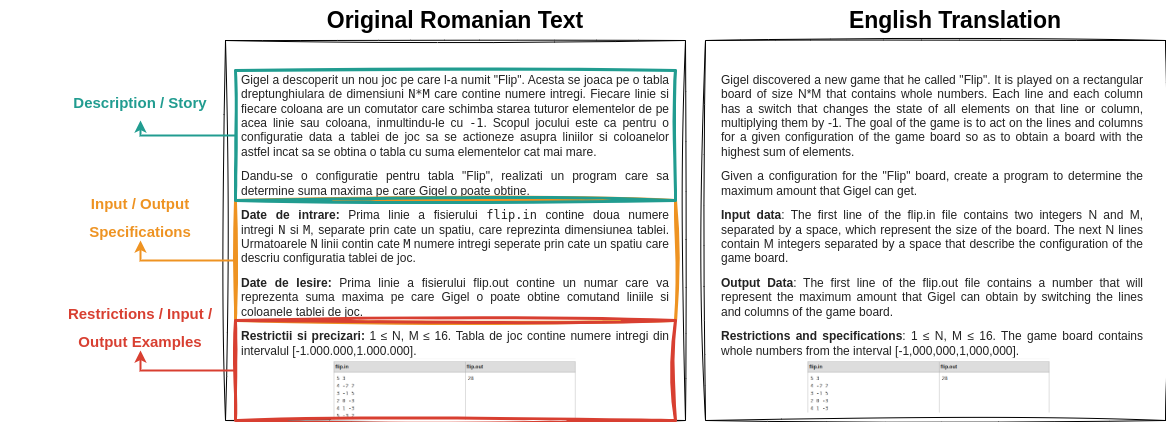}
    \caption{Example Romanian problem statement from RoCode. Problems specify a story-like description followed by input and output data specifications and restrictions. Finally, an example of an output from a correct solution is provided. We also show the English translation for non-Romanian readers.}
    \label{fig:example-statement}
\end{figure*}

InfoArena is a ``wiki'', in which volunteers can submit problems and discuss about solutions. The website is primarily addressed to Romanian students, and all problem statements are written in native Romanian. Problems usually follow a common format, containing an initial preamble, providing context, followed by the requirements and input/output data specification. Most problems also have concrete examples for the input and output data for a correct solution. However, not all hosted problems are valid. Out of the 3,072 problems, we filtered out problems that have no submitted solutions, contain no examples, or do not follow the suggested problem template. After filtering, we obtain a total of 2,642 problem statements. Originally, each problem statement was written using markdown. We cleaned any markdown formatting and left only the problem text in the same format for all problems. Figure \ref{fig:example-statement} showcases the common problem format in RoCode. In all problems, input data is given in a file and programs must output the correct solution in an output file. This is different from, for instance, APPS \cite{hendrycksapps2021}, in which input and output are served from standard input / output. 

\subsection{Solutions}
Each problem contains an average of $260.92$ solutions, with easier problems such as ``greatest common divisor'' containing the most submissions, while Olympiad-level problems contain just a few solutions. Originally, solutions were written in C / C++ or Pascal. We kept only C / C++ solutions, removed redundant comments and formatted every solution to a common code style using a public tool\footnote{\url{https://github.com/dawnbeen/c_formatter_42}}. We provide all $653,094$ solutions, and also a curated set of $11k$ solutions, which correspond to the top-$3$ highest-scoring, shortest solutions for each problem. Moreover, for the same author, we filtered similar solutions that obtained the same score for a source problem using the standard Levenshtein distance.

\begin{table}[hbt!]
    \centering
    \includegraphics[width=\linewidth]{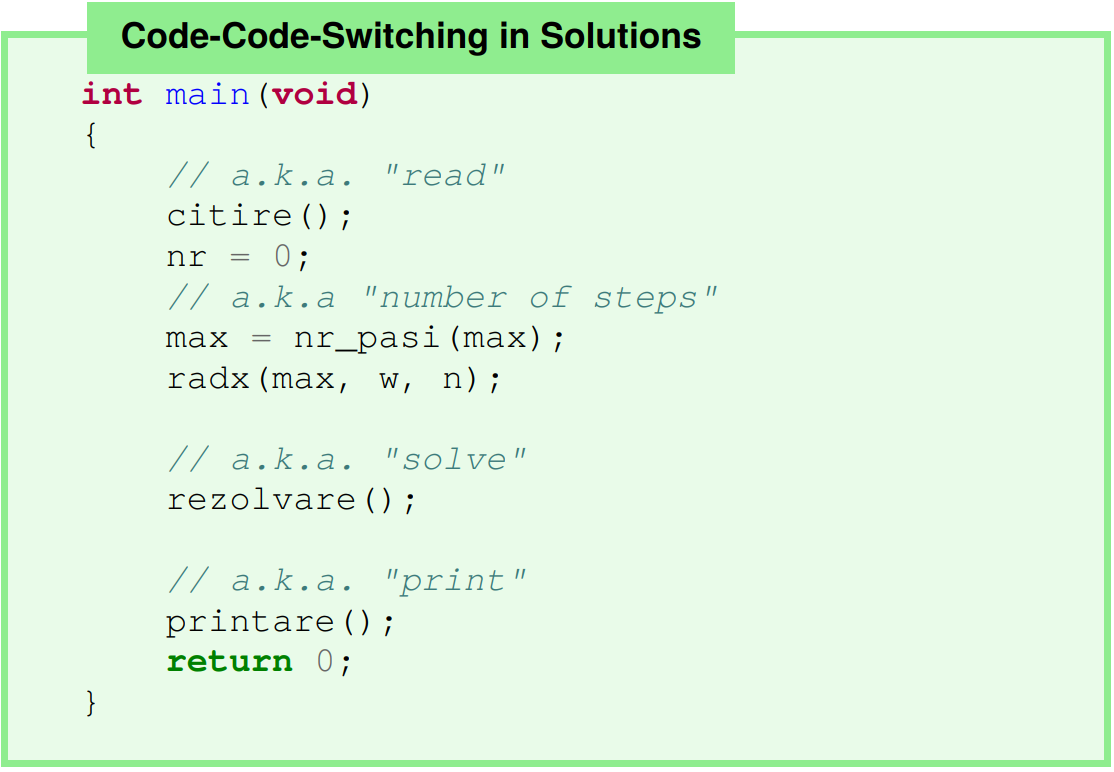}
    \caption{Example of a snippet from a C / C++ solution that exhibits code-code-switching: function names / variables are written in Romanian, whereas language keywords are written in English. Comments added by the paper authors.}
    \label{tab:code-code-switching}
\end{table}

Other datasets \cite{hendrycksapps2021,chen2021evaluating} and code-generation models \cite{husain2019codesearchnet,chen2021evaluating} are less focused on low-level C / C++, but more on high-level languages such as PHP, JavaScript and Python. Since Python has gained massive popularity in recent years, and is considered the de facto standard high-level programming language for machine learning, we also provide transpiled solutions in Python, using the OpenAI API. We used the ``\textit{gpt-3.5-turbo}'' model variant, which is a language model based on InstructGPT \cite{ouyang2022training}. We used the following prompt to transpile solutions: ``\textit{Translate the following C / C++ code to Python. Output only code, without any explanation or comments. Omit type hinting in the Python code: \{code\}}''. Transpiled solutions are automatically checked for correctness. Table \ref{tab:transpilation} showcases an example of a solution automatically transpiled into Python. The transpiled solution is more concise and preserves functionality.
\input{table/transpiled}

\begin{figure}
    \centering
    \includegraphics[width=0.85\linewidth]{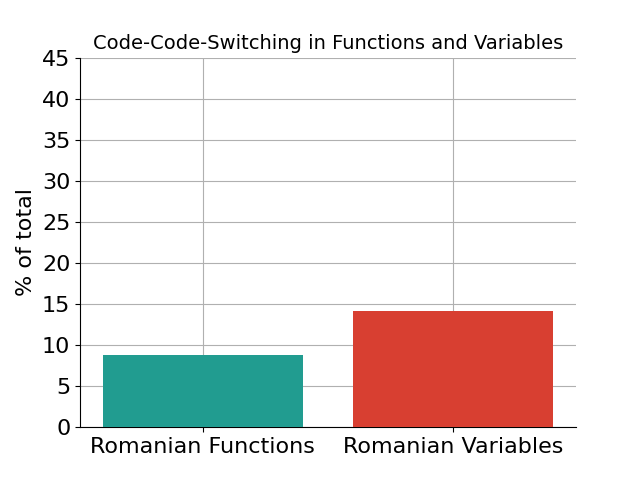}
    \caption{Percentage of Romanian variable names and function names that contain explicit Romanian words in submitted \textbf{human} C / C++ solutions. Abbreviations are omitted, making the actual proportion larger than shown here.}
    \label{fig:codes}
\end{figure}

Different from other English-oriented datasets, RoCode contains code that exhibits \textit{code-code-switching}: some function and variable names are written in Romanian, while others (e.g. language-specific keywords) are written in English. Table \ref{tab:code-code-switching} showcases a real snippet found in RoCode that has all function names written in Romanian, or abbreviated forms of Romanian words\footnote{``\textit{nr}'' is the abbreviation for ``\textit{num\u{a}r}'', equivalent to ``\textit{no.}'' for ``\textit{number}'' in English}. For instance, some problem solvers write ``\textit{rezolva()}'' instead of ``\textit{solve()}'', ``\textit{afisare()}'' instead of ``\textit{print()}''. This aspect provides additional complexity for adapting pretrained models to RoCode solutions. In Figure \ref{fig:codes}, we show the proportion of variable names and function names that contain Romanian words. We parsed the abstract syntax tree using ClangCheck LLVM \cite{LLVM} and uniformized all declarations to snake\_case format, since programmers use both camelCase, snake\_case or PascalCase. If any of the strings separated by underscore is found to be Romanian, we count that name as a Romanian variable / function. We used the Romanian WordNet (RoWordNet) \cite{dumitrescu2018rowordnet} to check if a word belongs to the Romanian language. We obtained that around 9\% of function names and around 14\% of variable names have explicit Romanian words. This approach counts only properly written words with only a subset of declensions, and omits abbreviations, which makes the actual counts higher than shown here. Examples of composite function names are: \textit{"acopera\_tot()" (cover\_everything()), "descompunereNumar()" (decomposeNumber()), "RezolvareDistantaMinimaCoborare()" (SolveMinimalDistanceDown())}. Examples of composite variable names are: \textit{"AdaugaValoare" (AddValue), "viziteazaTraseu" (visitRoute), deja\_castigat (already\_won)}.

\subsection{Test Cases}
Each solution is accompanied by a series of tests which are used to measure the correctness and computational complexity of the provided solution. The tests consist of an input file containing input data and an output file containing the desired output. There are a total of $35,758$ tests, and each problem has an average of $13$ tests, while some problems have upwards of $100$ tests. Since some tests are upwards of 100MB, we provide the smallest 5 tests for each problem, as well as an environment that automatically scores the provided generated solution. Similar to other works, problems are graded using accuracy, strict accuracy and  "pass@k" metric \cite{chen2021evaluating}.

\subsection{Estimating Problem Difficulty}
\begin{figure}[hbt!]
    \centering
    \includegraphics[width=0.75\linewidth]{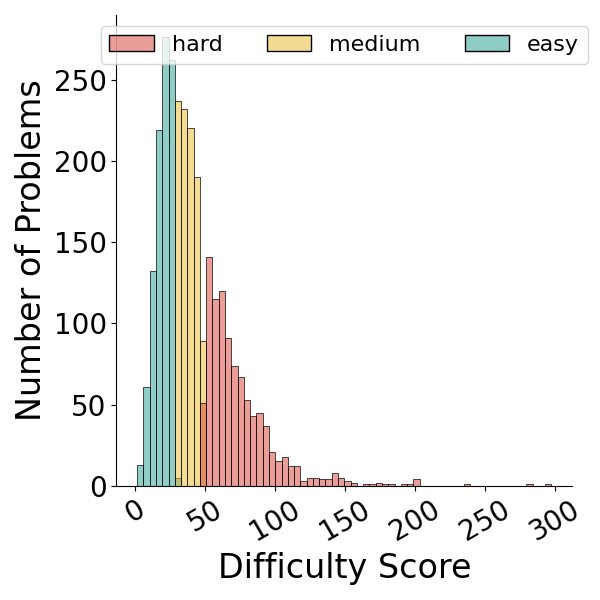}
    \caption{Distribution of problem difficulties in RoCode. Problem difficulty is estimated automatically based on the number of correct solutions, unique users, and the date of the problem.}
    \label{fig:distribution-diff}
\end{figure}

\begin{figure}[hbt!]
    \centering
    \includegraphics[width=0.75\linewidth]{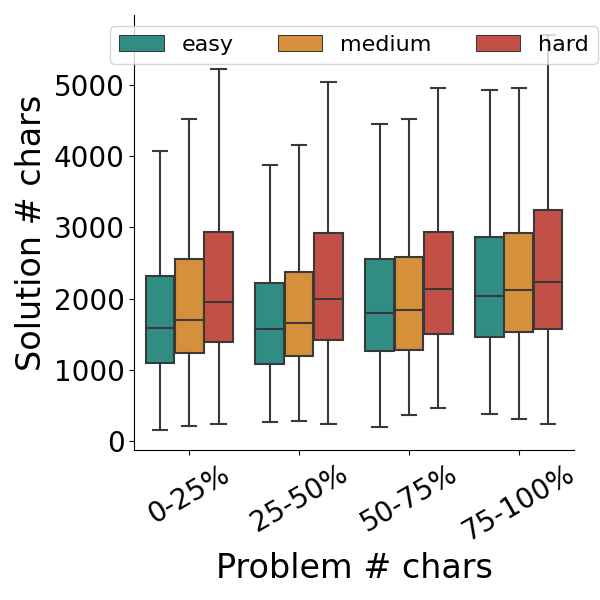}
    \caption{Correlation between problem statement length and solution lengths across problem difficulties. Harder problems require longer solutions, and problem statement length is not correlated with solution length.}
    \label{fig:len-len}
\end{figure}

Following similar code datasets \cite{hendrycksapps2021,chen2021evaluating} which provide different difficulty splits, we compute a difficulty score for each problem. For each problem, we computed the average score for the submitted user solutions, divided by the number of unique users. Since the publication date for problems ranges from the year 2006 up to 2022, we divided this score by the recency. We subsequently split RoCode into "easy", "medium" and "hard" problems, by splitting the difficulty distribution into tertiles. Consequently, we obtain a total of 790, 922 and 934 easy, medium and hard problems, respectively. Figure \ref{fig:distribution-diff} showcases the distribution of problem difficulties across RoCode. Furthermore, Figure \ref{fig:len-len} showcases the correlation between problem lengths and average solution lengths for each difficulty label - we found no significant correlation between lengths of problems and solutions, but consistently harder problems require longer solutions. Regarding \textit{code-code-switching}, perhaps surprisingly, we found the same distribution of Romanian variable names and function names across difficulty labels. 

\subsection{Dataset Splits}
We split the dataset into training, validation and test subsets, to enable researchers to fine-tune code generation models and to also provide a common testing split to compare approaches. The training, validation and test splits contain $2112$, $264$ and $266$ problems, respectively. We sampled problems uniformly across problem difficulty for each data split.

%% file: table/transpiled.tex
\begin{table*}[hbt]
    \centering
\begin{tabular}{p{0.5\linewidth} p{0.5\linewidth}}
Original C / C++ Solution & Transpiled into Python \\
\midrule
\begin{minipage}[t]{0.5\linewidth}
{\tiny
\begin{minted}{c}
#include <bits/stdc++.h>
#define all(cont) cont.begin(), cont.end()
#define pb push_back
using namespace std;
ifstream f("benzina.in");
ofstream g("benzina.out");
int main(void)
{
	int n; long long ans = 0;
	f >> n; n *= 2;
	vector<int> a(n);
	priority_queue<int> heap;
	for (int i = 0; i < n; ++i)
		f >> a[i];
	for (int i = 0; i < n; ++i)
	{ int cost; f >> cost; ans += cost; a[i] -= cost; }
	for (int i = 0; i < n; ++i)
	{
		heap.push(a[i]);
		if (i % 2 == 0)
		{ ans += heap.top(); heap.pop(); }
	}
	g << ans << '\n';
	f.close(); g.close();
}
\end{minted}
}
\end{minipage}
& 
\begin{minipage}[t]{0.5\linewidth}
{\tiny
\begin{minted}{python}
import heapq
with open('benzina.in') as f, open('benzina.out', 'w') as g:
    n = int(f.readline().strip())
    n *= 2
    a = [int(x) for x in f.readline().split()]
    ans = 0
    for i in range(n):
        cost = int(f.readline())
        ans += cost
        a[i] -= cost
    heap = []
    for x in a:
        heapq.heappush(heap, x)
        if len(heap) % 2 == 1:
            continue
        ans += heapq.heappop(heap)
    g.write(str(ans) + '\n')
\end{minted}
}
\end{minipage}
\end{tabular}
    \caption{Example of transpiled solution into Python, using ``\textit{gpt-3.5-turbo}''. The Python code preserves functionality, while being comparatively smaller. For the C / C++ version, we formatted the code to be more concise for display purposes.}
    \label{tab:transpilation}
\end{table*}

%% file: sections/4.results.tex
\subsection{Experimental Setup}
For benchmarking existing language models on RoCode, we follow the evaluation procedure proposed by \citet{hendrycksapps2021}. We used the following prompt\footnote{English Translation: "The following programming problem is given. The code in Python3 that solves the problem,
without comments or explanations, is:" } for all models:

\noindent \texttt{Se dă urmatoarea problemă de programare:\\
<PROBLEM STATEMENT>\\
<INPUT / OUTPUT SPEFICIATIONS>\\
<EXAMPLES>\\
Codul în Python3 care rezolvă problema, fără comentarii sau explicații, este:
}

In our experiments, we generated 10 solutions per problem. We computed accuracy, strict accuracy and pass@k \cite{chen2021evaluating} metrics. For accuracy, we counted the average maximum number of tests passed per problem. For strict accuracy, we counted the average number of times a problem has passed all tests. Additionally, we used the "pass@k" metric for evaluation. We generate 10 samples per problem and count the number of correct samples c. In particular, due to computational constraints, we chose $k \in \{1, 10\}$. The metric is defined as:

\begin{equation}
    \verb|pass@k| := \mathbb{E} \left[ 1 - \frac{\binom{n - c}{k}}{\binom{n}{k}}\right]
\end{equation}

All models in this work were executed with a temperature of 0.2 and top-p sampling of 0.90. We used a timeout of 4 seconds for solution run time. Models are run on a machine with 2x NVIDIA RTX 3060 with 12GB of VRAM each. Bigger models that did not fit a single GPU were parallelized across the two GPUs. All models' performance is measured in zero-shot settings (no fine-tuning, no additional examples or solution peaking \cite{hendrycksapps2021}). We used publicly available models from HuggingFace in all instances. In case the generated solution does not write to a file (as required by the problem statement), and instead expects the input and output to be managed through standard input and output, we obliged and provided the test data accordingly.

\subsection{Benchmarking Romanian Language Models}

We evaluated existing Romanian language models: Ro-GPT2 \cite{9643330} in three model sizes (124M, 354M and 774M parameters) and GPT-Neo-Ro \cite{ro-transformers}. Unsurprisingly, \textbf{None} of the currently available Romanian language models are able to understand the problem definition or to produce code, and all generated code could not be compiled. Performance in terms of accuracy, strict accuracy, pass@1 and pass@10 is exactly 0 for all metrics. This performance can be partially explained by model size, as similar English-based models (e.g. GPT-Neo \cite{pile}) also have below 3\% pass rate on easy English problems \cite{hendrycksapps2021}, but have a comparatively larger pretraining dataset. We point out that RoGPT-2’s corpus (for example) contains around ~3,400M tokens, but contains little to no code tokens, which explains the poor performance. We provide a more detailed discussion below.

\subsection{Benchmarking English Models}
In Table \ref{tab:en-results}, we showcased results for 4 open-source models with a relatively small number of parameters (maximum $7B$). We chose a selection of small and efficient models due to computational limitations. We leave more extensive evaluations, as well as different fine-tuning schemes \cite{hu2021lora}, as future work. Similar to the Romanian models, all models' performance is measured in zero-shot settings. We evaluated replit-code-v1-3b \cite{replit-code}, a small but capable model trained on the Stack 3T \cite{kocetkov2022stack}, that outperforms bigger models on code intelligence tasks. Furthermore, we also evaluated LLaMA-7b \cite{touvron2023llama} and OpenLLaMA-7b \cite{openlm2023openllama}, which are trained on 1T and 300B tokens, respectively, of both English and code. Finally, we evaluated WizardCoder-Python-7b \cite{luo2023wizardcoder}, an instruction fine-tuned variant of LLaMA-7b, through automatically generated instructions, which showcases very good performance on code intelligence tasks, surpassing in some cases commercial models. English models we included in our study are pretrained on some Romanian data -- for example, for LLaMA-2 \cite{touvron2023llama}, the pretraining dataset contains 0.03\% Romanian tokens out of a total of 2T tokens ($\sim$ 600M Romanian tokens) -- a very small proportion of the pretraining data, smaller than, for instance RoGPT-2s's corpus, but not a small number of Romanian words in absolute terms. It is assumed that performance is improved by the model's ability to exploit cross-lingual commonalities. Evidently, the English-oriented models have poor performance, only solving a handful of easy problems. It is clear that more recent, larger models with code in the training set output plausible code solutions, which most of the time compile properly: on average, around 18\% of solutions result in compilation errors across models. In Table \ref{tab:example-outputs}, we showcased selected model outputs for 2 Romanian models and 2 English-oriented models. We further provide some insights to the unsatisfactory performance of existing models.

\input{table/english-results}

\input{table/llm-output}
\subsection{Discussion}

There are several reasons for the bad performance of Romanian models, which also provide future research directions for specialized monolingual and general-purpose models on languages other than English:  

\textbf{No code data present in the pre-training dataset.} Romanian models have been trained on datasets derived from OSCAR corpus \cite{oscar}, Wikipedia and books, and do not have dedicated code splits annotated with Romanian text. Even if code is present in the dataset (from Common Crawl), it is described in English and comments and documentation are in English. The same argument can be made for mathematics and other scientific disciplines. A large, highly curated and dedicated Romanian dataset containing scientific data, mathematics and coding splits is needed.

\textbf{Current Romanian models are small and pre-training dataset is too noisy.} Small language models are prone to hallucinations and cannot follow instructions reliably without a high-quality curated dataset. Recently, models such as Mistral-7b \cite{jiang2023mistral} have shown that a relatively small model ($<7B$ parameters) can obtain comparable performance to much larger models by training for longer on a highly curated dataset. In the work by \citet{eldan2023tinystories}, the authors show that a small language model can still produce coherent text while only being trained on a small, easy to understand and curated dataset, raising questions whether model scale is the principal factor in model performance. Furthermore, the use of augmentation through retrieval (similar to RETRO \cite{borgeaud2021improving}) has been shown to increase model performance without increasing its size - such techniques have not been explored in monolingual models, for instance, retrieving tokens from translated English text during training.

\textbf{There is no post-training refinement for code.} Post-training techniques such as instruction-tuning \cite{mishra2022cross} is a proven method for better performance and controllability of LLM output by following natural language instructions. A Romanian dataset of instructions has not yet been compiled outside of automatically translated versions \cite{dac2023okapi}.

These negative results have not been addressed so far in the literature, and our hope is that it inspires future directions in training general-purpose Romanian or otherwise low-resourced language models. Further, we discuss the performance of English-oriented models.

\textbf{The pre-training set for English-oriented models might have data leakage from RoCode.} It is surprising that English-oriented models can somewhat follow the Romanian text description, essentially performing translation from Romanian text to Python code with symbols in English. However, it is unclear if correct model outputs can be attributed to proper text understanding or if it is a form of data leakage from larger corpora. By manually investigating the problems with high pass rate, the problems with the most tests passed are easy problems describing, for example, the edit distance, computing graph diameter (maximal distance between leafs), and detecting repeating sub-sequences. These are classic programming problems, and it is very likely that they appeared in other contexts in the pre-training sets, for instance in some parts of Common Crawl \cite{kudela2017extracting}, since the problem definitions are publicly available and are likely discussed on other websites. Moreover, as shown in Table \ref{tab:example-outputs} the output from \textit{open-llama-7b} is in C / C++ even though the prompt explicitly mentioned Python3. This is a further indication of data leakage. 

\begin{figure}[hbt!]
    \centering
    \includegraphics[width=0.75\linewidth]{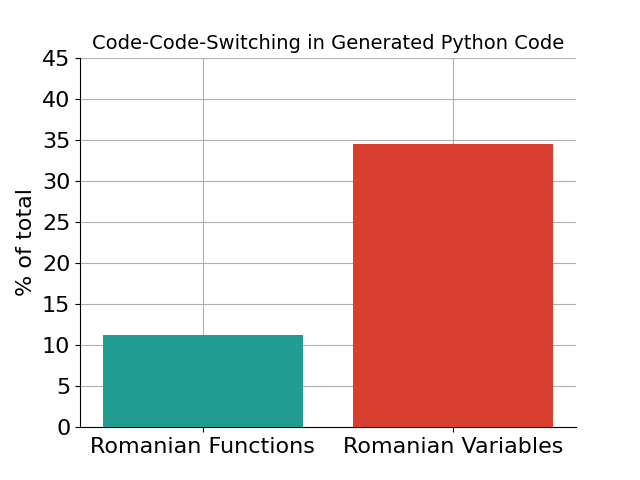}
    \caption{Percentage of Romanian variable and function names that contain explicit Romanian words in \textbf{model-generated} Python solutions. Generated solutions tend to have more explicit Romanian names.}
    \label{fig:code-code-switch-python}
\end{figure}

\textbf{English models exhibit even more \textit{code-code-switching} compared to human solutions.}  Furthermore, code generated from the language models exhibits even more \textit{code-code-switching} when generating Python code (see Figure \ref{fig:code-code-switch-python}). The models we tested tend to use many more Romanian variable names and slightly more Romanian functions compared to the human-submitted solutions (see Figure \ref{tab:code-code-switching}): we found around 35\% of variable names and 12\% of function names contain explicit Romanian words in the generated Python solutions. The larger amount of code-code-switching is presumably an artifact of the next token prediction objective, and the models usually follow the terminology present in the problem text and in the output examples. This leads the model to adopt a similar style of Romanian variable naming in further code blocks. 

\textbf{Translating the problem definitions in English worsened results.} Evaluating a \textit{replit-v1-3b} \cite{replit-code} on translated problem definitions resulted in 1.26\% accuracy, down from 1.81\% using original Romanian descriptions. This is due both to imprecise translation and obfuscating exact formulations that might be present in the pretraining set. Translation is not a long-term solution to multilingual code intelligence models and it does not substitute proper language and code understanding of specialized monolingual models.

%% file: table/english-results.tex
\begin{table}[hbt!]
    \centering
    \resizebox{\linewidth}{!}{
    \begin{tabular}{lc|cc|cc}
        \textbf{Model} & \textbf{\# Params} & \textbf{Acc. \%} &  \textbf{Strict Acc. \%} & \textbf{pass@1} & \textbf{pass@10}  \\
        \midrule
        llama-2-7b      & 7B & 0.55 & 0.0 & 0.02 & 0.98 \\
        open-llama-2-7b & 7B & 1.50 & 0.0 & 0.05 & 2.37 \\
        replit-code-v1-3b & 3B & 1.81 & 0.39 & 0.1 & 7.74 \\
        wizard-coder-7b & 7B & 3.86 & 0.79 & 0.27 & 3.83 \\
    \end{tabular}}
    \caption{Performance on RoCode for open-sourced English-oriented code models. While performance is different then zero, current models struggle to solve the problems expressed in Romanian.}
    \label{tab:en-results}
\end{table}

%% file: table/llm-output.tex
\begin{table*}[hbt!]
    \centering
\begin{tabular}{p{0.15\linewidth} p{0.15\linewidth} p{0.23\linewidth} p{0.23\linewidth}}
RoGPT2-large \cite{9643330} & GPT-Neo-Ro \cite{ro-transformers} & replit-code-v1-3b \cite{replit-code} & Open-LLaMa-7b \cite{openlm2023openllama} \\
\midrule
{\tiny
\textit{"Votul este acordat(x) din X/C(n).Se presupune ca ambii(X,n) este numărul de X\% pentru care Tony are dreptul\%(y,j,k,o) voturi de votat.JPGPunctul (y) xq voturi din 1/3.pdfqq.Este numarulq voturile Xq pe votul fanilor ingustequale. Poses(q) Q(sq(t) voturile de votati inqual pentru votul lui.qualitatiunilor pentru al X(z(Z(k)z-ului).pdfPseud.p(m). \dots \dots"}
}
& 
{\tiny
\textit{"Python 3 Rezolvă această problemă, nu sunt necesare nici un fel de comentarii. Doar faceți click pe această pagină, și verificați codul. Există și o variantă pentru limba română: Pyyp4. În ea se scrie următorul cod: -y
Un exemplu clasic este prezentat în continuare. O mașină de măsurat cu trei axe dispune de un sistem de operare care rulează într-o bază de date, numit "TANNO", în care pot exista fișiere cu datele măsurate de cele trei modele ale mașinii \dots \dots"}
}
&
\begin{minipage}[t]{0.25\linewidth}
{\tiny
\begin{minted}{python}
n = int(input())
candidates = list(map(
    str, input().split())
)
candidates.sort()
votes = list(map(
    int, 
    input().split())
)
votes.sort()
Q = int(input())
for q in range(Q):
    L, R = map(int, input().split())
    print(votes.count(L) +\ 
        votes.count(R))
\end{minted}
}
\end{minipage}
&
\begin{minipage}[t]{0.25\linewidth}
{\tiny
\begin{minted}{cpp}
#include <stdio.h>
#include <stdlib.h>
#define MAX_N 1000000
#define MAX_S 1000000
int main()
{
    int n, q;
    scanf("%d %d", &n, &q);
    char *votes[MAX_N];
    for (int i = 0; i < n; i++){
        votes[i] = (char *)malloc(MAX_S);
        scanf("%s", votes[i]);}
    int ans = 0;
    for (int i = 0; i < q; i++)
    {
        int l, r;
        scanf("%d %d", &l, &r);
        int cnt = 0;
        for (int j = l; j <= r; j++)
            if (votes[j][0] == 'C')
                cnt++;
        if (cnt < r - l + 1){
            for (int j = l; j <= r; j++)
                if (votes[j][0] == 'C')
                    votes[j][0] = 'T';
            ans++;}
    }
    printf("%d\n", ans);
    return 0;
}

\end{minted}
}
\end{minipage}
\end{tabular}
    \caption{Selected outputs of open-source language models for an easy problem from RoCode. Romanian models are not able to output code, and instead hallucinate text vaguely similar to the problem description. We truncated the output for display purposes, as the quality degrades severely. The \textit{replit-code-v1-3b} model outputs plausible Python3 code, but it does not correctly address the problem. \textit{Open-LLaMa-7b} outputs C / C++ code instead of the required Python.}
    \label{tab:example-outputs}
\end{table*}

%% file: sections/5.conclusions.tex
In this work, we presented RoCode, a benchmarking dataset for code intelligence systems that measures the understanding of problem definitions in Romanian to provide algorithms that correctly solve the problems. Competitive programming benchmarks \cite{hendrycksapps2021} are still a challenging task, even for current state-of-the-art commercial models. However, all training sets containing code and benchmarks are implicitly geared towards English, with documentation, comments and problem definitions written solely in English. RoCode fills the gap in the benchmarking suite for Romanian NLP systems such as LiRo \cite{liro2021}, which do not have any tasks for code generation for Romanian. Our dataset is challenging: several Romanian and English-oriented language models that we tested have poor performance, managing to correctly solve only a handful of problems from the test set. This work paves the way for further research of large language models for code intelligence in non-English languages and is a preliminary step in the democratization of programming for non-English speakers.

%% file: sections/6.ack.tex
We thank the \textit{infoarena.ro} team for providing the raw data for RoCode. The work of Adrian Cosma was performed as part of Short-Term Research Mission (STSM) at Universitat Politècnica de València, part of COST Action CA18231, Multi3Generation: Multi-task, Multilingual, Multi-modal Language Generation. The work of Paolo Rosso was in the framework of the FairTransNLP research project (PID2021-124361OB-C31), funded by MCIN/AEI/10.13039/501100011033 and by ERDF, EU A way of making Europe. 